\documentclass[11pt,letterpaper]{article}
\usepackage{emnlp2016}
\usepackage{times}
\usepackage{latexsym}
\usepackage{lipsum}
\usepackage{booktabs}
\usepackage{subfig}

\newif\ifcomment
\commenttrue

\newif\ifanonymous
\anonymousfalse

\def \eg {\emph{e.g.},}
\def \ie {\emph{i.e.},}

\usepackage{color}

\usepackage[colorinlistoftodos]{todonotes}

\ifcomment
\newcommand{\thms}[1]{\todo[inline,color=orange!75]{#1}}
\newcommand{\tim}[1]{\todo[inline,color=blue!10]{#1}}
\newcommand{\seb}[1]{\todo[inline,color=yellow!60]{#1}}

\else
\newcommand{\thms}[1]{}
\newcommand{\tim}[1]{}
\newcommand{\seb}[1]{}

\fi

\ifanonymous
\def \akbc{\newcite{demeester2016akbc-anon}}
\def \akbcb{\cite{demeester2016akbc-anon}}
\else
\def \akbc{\newcite{demeester2016akbc}}
\def \akbcb{\cite{demeester2016akbc}}
\fi

\usepackage{bm}
\newcommand{\emb}[1]{\bm{#1}}
\newcommand{\rel}[1]{\bm{#1}}
\newcommand{\ent}[1]{\bm{#1}}
\newcommand{\fact}[2]{\langle #1,#2\rangle}
\newcommand{\dotre}[2]{\rel{#1}^\top\ent{#2}}

\def \R {\mathcal{R}}
\def \T {\mathcal{T}}

\usepackage{amsmath}
\usepackage{amssymb}

\def \Real {\mathbb{R}}
\def \cleq {\leq} 
\def \cgeq {\geq}

\usepackage{url}

\ifanonymous
\else
\emnlpfinalcopy
\fi

\def\emnlppaperid{861}

\title{Lifted Rule Injection for Relation Embeddings}

\author{Thomas Demeester \\
  Ghent University - iMinds\\
  Ghent, Belgium \\
  {\small\tt tdmeeste@intec.ugent.be} \\\And
  Tim Rockt\"aschel \and Sebastian Riedel \\
  University College London \\
  London, UK \\
  {\small\tt \{t.rocktaschel,s.riedel\}@cs.ucl.ac.uk} \\}

\date{}

\begin{document}

\maketitle

\begin{abstract}
Methods based on representation learning currently hold the state-of-the-art in many natural language processing and knowledge base inference tasks.
Yet, a major challenge is how to efficiently incorporate commonsense knowledge into such models.
A recent approach regularizes relation and entity representations by propositionalization of first-order logic rules.
However, propositionalization does not scale beyond domains with only few entities and rules.
In this paper we present a highly efficient method for incorporating implication rules into distributed representations for automated knowledge base construction.
We map entity-tuple embeddings into an approximately Boolean space and encourage a partial ordering over relation embeddings based on implication rules mined from WordNet.
Surprisingly, we find that the strong restriction of the entity-tuple embedding space does not hurt the expressiveness of the model and even acts as a regularizer that improves generalization.
By incorporating few commonsense rules, we achieve an increase of $2$ percentage points mean average precision over a matrix factorization baseline, while observing a  negligible increase in runtime.
\end{abstract}

\section{Introduction}

Current successful methods for automated knowledge base construction tasks heavily rely on learned distributed vector representations \cite{nickel2012factorizing,riedel2013relation,socher2013reasoning,chang2014typed,neelakantan2015compositional,toutanova2015representing,nickel2015review,Verga2016a,Verga2016b}.
Although these models are able to learn robust representations from large amounts of data, they often lack commonsense knowledge. 
Such knowledge is rarely explicitly stated in texts but can be found in resources like PPDB \cite{ganitkevitch2013ppdb} or WordNet \cite{miller1995wordnet}.

Combining neural methods with symbolic commonsense knowledge, for instance in the form of implication rules, is in the focus of current research \cite{rocktaschel2014low,wang2014structure,bowman2015recursive,wang2015kbc,vendrov2016orderembeddings,hu2016harnessing,rocktaschel2016learning,Cohen2016TensorLog}.
A recent approach \cite{rocktaschel2015injecting} regularizes entity-tuple and relation embeddings via first-order logic rules.
To this end, every first-order rule is propositionalized based on observed entity-tuples, and a differentiable loss term is added for every propositional rule.
This approach does not scale beyond only a few entity-tuples and rules. For example, propositionalizing the rule $\forall x: \verb~isMan~(x) \Rightarrow \verb~isMortal~(x)$ would result in
a very large number of loss terms on a large database.

In this paper, we present a method to incorporate simple rules while maintaining the computational efficiency of only modeling training facts.
This is achieved by minimizing an upper bound of the loss that encourages the implication between relations to hold, entirely independent from the number of entity pairs.
It only involves representations of the relations that are mentioned in rules, as well as a general rule-independent constraint on the entity-tuple embedding space. 
In the example given above, if we require that every component of the vector representation of {\verb isMan } is smaller than the corresponding component of relation {\verb isMortal }, then we can show that the rule holds for any \emph{non-negative} representation of an entity-tuple.
Hence our method avoids the need for separate loss terms for every ground atom resulting from propositionalizing rules. 
In statistical relational learning this type of approach is often referred to as \emph{lifted} inference or learning~\cite{Poole:2003:FPI:1630659.1630801,Braz:2007:LFP:1369181} because it deals with groups of random variables at a first-order level. 
In this sense our approach is a lifted form of rule injection. 
This allows for imposing large numbers of rules while learning distributed representations of relations and entity-tuples.
Besides drastically lower computation time, an important advantage of our method over \newcite{rocktaschel2015injecting} is that when these constraints are satisfied, the injected rules always hold, even for unseen but inferred facts.
While the method presented here only deals with implications and not general first-order rules, it does not rely on the assumption of independence between relations, and is hence more generally applicable.

Our contributions are fourfold: 
(i) we develop a very efficient way of regularizing relation representations to incorporate first-order logic implications (\S\ref{sec:model}),
(ii) we reveal that, against expectation, mapping entity-tuple embeddings to non-negative space does not hurt but instead improves the generalization ability of our model (\S\ref{subsec:Frestricted})
(iii) we show improvements on a knowledge base completion task by injecting mined commonsense rules from WordNet (\S\ref{subsec:injectWordNet}), 
and finally (iv) we give a qualitative analysis of the results, demonstrating that implication constraints are indeed satisfied in an asymmetric way and result in a substantially increased structuring of the relation embedding space (\S\ref{subsec:asymm}).

\section{Background}\label{sec:background}

In this section we revisit the matrix factorization relation extraction model by \newcite{riedel2013relation} and introduce the notation used throughout the paper.
We choose the matrix factorization model for its simplicity as the base on which we develop implication injection. 

\newcite{riedel2013relation} represent every relation $r\in\R$ (selected from Freebase \cite{bollacker2008freebase} or extracted as textual surface pattern) 
by a $k$-dimensional latent representation $\rel{r}\in\Real^k$. 
A particular \emph{relation instance} or \emph{fact} is the combination of a relation $r$ and a tuple $t$ of entities that are engaged in that relation, and is written as $\fact{r}{t}$. 
We write $\mathcal{O}$ as the set of all such input facts available for training.
Furthermore, every entity-tuple $t\in\mathcal{T}$ is represented by a latent vector $\ent{t}\in\Real^k$ (with $\mathcal{T}$ the set of all entity-tuples in $\mathcal{O}$).

Model F by \newcite{riedel2013relation} measures the compatibility between a relation $r$ and an entity-tuple $t$ using the dot product $\dotre{r}{t}$ of their respective vector representations. 
During training, the representations are learned such that valid facts receive high scores, whereas negative ones receive low scores. 
Typically no negative evidence is available at training time, and therefore a Bayesian Personalized Ranking (BPR) objective~\cite{Rendle2009BPR} is used.
Given a pair of facts $f_p:=\fact{r_p}{t_p}\not\in\mathcal{O}$ and $f_q:=\fact{r_q}{t_q}\in\mathcal{O}$, this objective requires that
\begin{equation}
\dotre{r_p}{t_p}\leq\dotre{r_q}{t_q}.
\end{equation}
The embeddings can be trained by minimizing a convex loss function $\ell_R$ that penalizes violations of that requirement when iterating over the training set. 
In practice, each positive training fact $\fact{r}{t_q}$ is compared with a randomly sampled unobserved fact $\fact{r}{t_p}$ for the same relation. 
The overall loss can hence be written as
\begin{equation}
\mathcal{L}_R = \sum_{\substack{\fact{r}{t_q}\in\mathcal{O}\\t_p\in\mathcal{T},\  \fact{r}{t_p}\not\in\mathcal{O}}} \ell_R\big(\emb{r}^\top[\emb{t_p}-\emb{t_q}]\big).\label{eq:LR}
\end{equation}
and measures how well observed valid facts are ranked above unobserved facts, thus reconstructing the ranking of the training data.
We will henceforth call $\mathcal{L}_R$ the \emph{reconstruction loss}, to make a distinction with the \emph{implication loss} that we will introduce later. 
\newcite{riedel2013relation} use the logistic loss $\ell_R(s):=-\log\sigma(-s)$,
where $\sigma(s):=(1+e^{-x})^{-1}$ denotes the sigmoid function. 
In order to avoid overfitting, an $L_2$ regularization term on the $\emb{r}$ and $\emb{t}$ embeddings is added to the reconstruction loss.
The overall objective to minimize hence is
\begin{equation}
\mathcal{L}_F = \mathcal{L}_R + \alpha \big({\textstyle\sum_{\emb{r}}}\|\emb{r}\|_2^2+{\textstyle\sum_{\emb{t}}}\|\emb{t}\|_2^2\big)
\end{equation}
where $\alpha$ is the regularization strength.

\section{Lifted Injection of Implications}\label{sec:model}

In this section, we show how an implication
\begin{equation}
\forall t\in\T: \fact{r_p}{t}\Rightarrow\fact{r_q}{t},\label{eq:def_impl}
\end{equation}
can be imposed independently of the entity-tuples. 
For simplicity, we abbreviate such implications as $r_p\Rightarrow r_q$ (\eg{} $\text{\tt{professorAt}}\Rightarrow\text{\tt{employeeAt}}$).

\subsection{Grounded Loss Formulation}
The implication rule can be imposed by requiring that every tuple $t\in\T$ is at least as compatible with relation $r_p$ as with $r_q$. 
Written in terms of the latent representations, eq.~\eqref{eq:def_impl} therefore becomes
\begin{equation}
\forall t\in\T: \dotre{r_p}{t}\leq\dotre{r_q}{t}\label{eq:def2_impl}
\end{equation}
If $\fact{r_p}{t}$ is a true fact with a high score $\dotre{r_p}{t}$, and the fact $\fact{r_q}{t}$ has an even higher score, it must also be true, but not vice versa. 
We can therefore inject an implication rule by minimizing a loss term with a separate contribution from every $t\in\mathcal{T}$, adding up to the total loss if the corresponding inequality is not satisfied. 
In order to make the contribution of every tuple~$t$ to that loss independent of the magnitude of the tuple embedding, 
we divide both sides of the above inequality by ${\|\emb{t}\|}_1$. With $\emb{\tilde{t}}:=\emb{t}/{\|\emb{t}\|}_1$, the implication loss for the rule $r_p\Rightarrow r_q$ can  be written as
\begin{equation}
\mathcal{L}_I= \sum_{\forall t\in\mathcal{T}} 
\ell_I\big([\emb{r_p}-\emb{r_q}]^\top\emb{\tilde{t}}\big)
\label{eq:loss1_impl}
\end{equation}
for an appropriate convex loss function~$\ell_I$, similarly to eq.~(\ref{eq:LR}).
In practice, the summation can be reduced to those tuples that occur in combination with $r_p$ or $r_q$ in the training data. Still, the propositionalization in terms of training facts leads to a heavy computational cost for imposing a single implication, similar to the technique introduced in \newcite{rocktaschel2015injecting}.
Moreover, with that simplification there is no guarantee that the implication between both relations would generalize towards inferred facts not seen during training.

\subsection{Lifted Loss Formulation}
The problems mentioned above can be avoided if instead of $\mathcal{L}_I$, a tuple-independent upper bound is minimized. Such a bound can be constructed, provided all components of $\emb{t}$ are restricted to a non-negative embedding space, \ie{} $\mathcal{T}\subseteq\Real^{k,+}$.
If this holds, Jensen's inequality allows us to transform eq.~(\ref{eq:loss1_impl}) as follows 
\begin{align}
\mathcal{L}_I 
&= \sum_{\forall t\in\mathcal{T}} \ell_I\bigg(\sum_{i=1}^k \tilde{t}_i\ [\emb{r_p}-\emb{r_q}]^\top\emb{1}_i\bigg)\\
&\leq \sum_{i=1}^k \ell_I\big([\emb{r_p}-\emb{r_q}]^\top\emb{1}_i\big) \sum_{\forall t\in\mathcal{T}}  \tilde{t}_i
\end{align}
where $\emb{1}_i$ is the unit vector along dimension $i$ in tuple-space. 
This is allowed because the $\{\tilde{t}_i\}_{i=1}^k$ form convex coefficients ($\tilde{t_i}>0$, and $\sum_i\tilde{t}_i=1$), and $\ell_I$ is a convex function.
If we define
\begin{equation}
\mathcal{L}_I^U := \sum_{i=1}^k  \ell_I\big([\emb{r_p}-\emb{r_q}]^\top\emb{1}_i\big) 
\end{equation}
we can write
\begin{equation}
\mathcal{L}_I\leq  \beta \mathcal{L}_I^U \label{eq:loss2_impl}
\end{equation}
in which $\beta$ is an upper bound on $\sum_t\tilde{t}_i$. 
One such bound is $|\mathcal{T}|$, but others are conceivable too. 
In practice we rescale $\beta$ to a hyper-parameter $\tilde{\beta}$ that we use to control the impact of the upper bound to the overall loss.
We call $\mathcal{L}_I^U$ the \emph{lifted loss}, as it no longer depends on any of the entity-tuples; it is grounded over the unit tuples $\emb{1}_i$ instead.

The implication $r_p\Rightarrow r_q$ can thus be imposed by minimizing the lifted loss $\mathcal{L}_I^U$. 
Note that by minimizing $\mathcal{L}_I^U$, the model is encouraged to satisfy the constraint $\rel{r_p}\cleq\rel{r_q}$ on the relation embeddings, where $\cleq$ denotes the component-wise comparison.
In fact, a sufficient condition for eq.~\eqref{eq:def2_impl} to hold, is 
\begin{equation}
\rel{r_p}\cleq\rel{r_q}\;\;\text{and}\;\; \forall t\in\T:\;\ent{t}\cgeq\emb{0}
\label{eq:obj3_impl}
\end{equation}
with $\emb{0}$ the $k$-dimensional null vector.
This corresponds to a single relation-specific loss term, and the general restriction $\mathcal{T}\subseteq\Real^{k,+}$ on the tuple-embedding space.

\subsection{Approximately Boolean Entity Tuples}\label{subsec:approx-bool}
In order to impose implications by minimizing a lifted loss $\mathcal{L}_I^U$, 
the tuple-embedding space needs to be restricted to $\Real^{k,+}$.
We have chosen to restrict the tuple space even more than required, namely to the hypercube $\emb{t}\in[0,1]^k$, as approximately Boolean embeddings \cite{kruszewski2015boolean}.
The tuple embeddings are constructed from real-valued vectors $\emb{e}$, using the component-wise sigmoid function
\begin{equation}
\emb{t} = \sigma(\emb{e}),\quad \emb{e}\in\Real^k.\label{eq:realvalued}
\end{equation}
For minimizing the loss, the gradients are hence computed with respect to $\emb{e}$, and the $L_2$ regularization is applied to the components of $\emb{e}$ instead of $\emb{t}$.

Other choices for ensuring the restriction $\emb{t}\cgeq\emb{0}$ in eq.~(\ref{eq:obj3_impl}) are possible, but we found that our approach works better in practice than those (\eg{} the exponential transformation proposed by \akbc). It can also be observed that the unit tuples over which the implication loss is grounded, form a special case of approximately Boolean embeddings.

In order to investigate the impact of this restriction even when not injecting any rules, we introduce model FS: the original model F, but with sigmoidal entity-tuples: 
\begin{align}
\mathcal{L}_{FS} =& 
\sum_{\substack{\fact{r}{t_q}\in\mathcal{O}\\t_p\in\mathcal{T},\  \fact{r}{t_p}\not\in\mathcal{O}}} \ell_R\big(\emb{r}^\top[\sigma(\emb{e_p})-\sigma(\emb{e_q})]\big) \nonumber\\
&+ \alpha \big({\textstyle\sum_{\emb{r}}}\|\emb{r}\|_2^2+{\textstyle\sum_{\emb{e}}}\|\emb{e}\|_2^2\big)
\end{align}
Here, $\emb{e_p}$ and $\emb{e_q}$ are the real-valued representations as in eq.~(\ref{eq:realvalued}), for tuples $t_p$ and $t_q$, respectively. 

With the above choice of a non-negative tuple-embedding space we can now state the full lifted rule injection model (FSL):
\begin{equation}
\mathcal{L}_{FSL} = \mathcal{L}_{FS} + \tilde{\beta} \sum_{I\in\mathcal{I}}\mathcal{L}^U_I
\end{equation}
$\mathcal{L}^U_I$ denotes a lifted loss term for every rule in a set $\mathcal{I}$ of implication rules that we want to inject.

\subsection{Convex Implication Loss}

The logistic loss $\ell_R$ (see \S\ref{sec:background}) is not suited for imposing implications because once the inequality in eq.~(\ref{eq:obj3_impl}) is satisfied, the components of $\emb{r_p}$ and $\emb{r_q}$ do not need to be separated any further. However, with $\ell_R$ this would continue to happen due to the small non-zero gradient. In the reconstruction loss $\mathcal{L}_R$ this is a desirable effect which further separates the scores for positive from negative examples.  
However, if an implication is imposed between two relations that are almost equivalent according to the training data, we still want to find almost equivalent embedding vectors. 
Hence, we propose to use the loss
\begin{equation}
\ell_I(s) = \max(0,s+\delta)
\end{equation}
with $\delta$ a small positive margin to ensure that the gradient does not disappear before the inequality is actually satisfied. We use $\delta=0.01$ in all experiments.

The main advantage of the presented approach over earlier methods that impose the rules in a grounded way 
\cite{rocktaschel2015injecting,wang2015kbc}
is the computational efficiency of imposing the lifted loss.
Evaluating $\mathcal{L}_I^U$ or its gradient for one implication rule is comparable to evaluating the reconstruction loss for one pair of training facts.
In typical applications there are much fewer rules than training facts and the extra computation time needed to inject these rules is therefore negligible.

\section{Related Work}\label{sec:relatedwork}

Recent research on combining rules with learned vector  representations has been important for new developments in the field of knowledge base completion.  
\newcite{rocktaschel2014low} and \newcite{rocktaschel2015injecting} provided a framework to jointly maximize the probability of observed facts and propositionalized first-order logic rules. 
\newcite{wang2015kbc} demonstrated how different types of rules can be incorporated using an Integer Linear Programming approach.  
\newcite{WangCohen2016matfact} learned embeddings for facts and first-order logic rules using matrix factorization.
Yet, all of these approaches ground the rules in the training data, limiting their scalability towards large rule sets and KBs with many entities. As argued in the introduction, this forms an important motivation for the lifted rule injection model put forward in this work, which by construction does not suffer from that limitation. 
\newcite{Wei2015} proposed an alternative strategy to tackle the scalability problem by reasoning on a filtered subset of grounded facts.

\newcite{WuAAAI2015} proposed to use a path ranking approach for capturing long-range interactions between entities, and to add these as an extra loss term, besides the loss that models pairwise relations. Our model FSL differs substantially from their approach, in that we consider tuples instead of separate entities, and we inject a given set of rules. Yet, by creating a partial ordering in the relation embeddings as a result of injecting implication rules, model FSL can also capture interactions beyond direct relations. 
This will be demonstrated in \S\ref{subsec:injectWordNet} by injecting rules between surface patterns only and still measuring an improvement on predictions for structured Freebase relations.

Combining logic and distributed representations is also an active field of research outside of automated knowledge base completion. Recent advances include the work by \newcite{faruqui2014retrofitting}, who injected ontological knowledge from WordNet into word representations. Furthermore, \newcite{vendrov2016orderembeddings} proposed to enforce a partial ordering in an embeddings space of images and phrases. Our method is related to such order embeddings since we define a partial ordering on relation embeddings. However, to ensure that implications hold for all entity-tuples we also need a restriction on the entity-tuple embedding space and derive bounds on the loss.
Another important contribution is the recent work by \newcite{hu2016harnessing}, who proposed a framework for injecting rules into general neural network architectures, by jointly training on the actual targets and on the rule-regularized predictions provided by a teacher network. Although quite different at first sight, their work could offer a way to use our model in various neural network architectures, by integrating the proposed lifted loss into the teacher network. 

This paper builds upon our previous workshop paper \akbcb.
In that work, we tested different tuple embedding transformations in an ad-hoc manner. We used approximately Boolean representations of relations instead of entity-tuples, strongly reducing the model's degrees of freedom. We now derive the FSL model from a carefully considered mathematical transformation of the grounded loss. The FSL model only restricts the tuple embedding space, whereby relation vectors remain real valued. 
Furthermore, previous experiments were performed on small-scale artificial datasets, whereas we now test on a real-world relation extraction benchmark.

Finally, we explicitly discuss the main differences with respect to the strongly related work from \newcite{rocktaschel2015injecting}.
Their method is more general, as they cover a wide range of first-order logic rules, whereas we only discuss implications. Lifted rule injection beyond implications will be studied in future research contributions. However, albeit less general, our model has a number of clear advantages:
\par \textbf{Scalability --}
Our proposed model of lifted rule injection scales according to the number of implication rules, instead of the number of rules times the number of observed facts for every relation present in a rule. 
\par \textbf{Generalizability --}
Injected implications will hold even for facts not seen during training, because their validity only depends on the order relation imposed on the relation representations. This is not guaranteed when training on rules grounded in training facts by \newcite{rocktaschel2015injecting}.
\par\textbf{Training Flexibility --}
Our method can be trained with various loss functions, including the rank-based loss as used in \newcite{riedel2013relation}.
This was not possible for the model of \newcite{rocktaschel2015injecting} and already leads to an improved accuracy as seen from the zero-shot learning experiment in \S\ref{subsec:zeroshot}. 
\par\textbf{Independence Assumption --}
In \newcite{rocktaschel2015injecting} an implication of the form $a_p\Rightarrow a_q$ for two ground atoms $a_p$ and $a_q$
is modeled by the logical equivalence $\neg(a_p \wedge \neg a_q)$, and its probability is approximated in terms of the elementary probabilities $\pi(a_p)$ and $\pi(a_q)$ as $1-\pi(a_p)\big(1-\pi(a_q)\big)$. 
This assumes the independence of the two atoms $a_p$ and $a_q$, which may not hold in practice.
Our approach does not rely on that assumption and also works for cases of statistical dependence.
For example, the independence assumption does not hold in the trivial case where the relations $r_p$ and $r_q$ in the two atoms are equivalent, whereas in our model, the constraints $\emb{r}_p\leq\emb{r}_q$ and  $\emb{r}_p\geq\emb{r}_q$ would simply reduce to $\emb{r}_p=\emb{r}_q$. 

\section{Experiments and Results}\label{sec:results}

We now present our experimental results. 
We start by describing the experimental setup and hyperparameters. 
Before turning to the injection of rules, we compare model F with model FS, and show that restricting the tuple embedding space has a regularization effect, rather than limiting the expressiveness of the model (\S\ref{subsec:Frestricted}).
We then demonstrate that model FSL is capable of zero-shot learning (\S\ref{subsec:zeroshot}), and show that injecting high-quality WordNet rules leads to an improved precision (\S\ref{subsec:injectWordNet}). 
We proceed with a visual illustration of the relation embeddings with and without injected rules (\S\ref{subsec:visualizing}), provide details on time efficiency of the lifted rule injection method (\S\ref{subsec:time}), and show that it correctly captures the asymmetry of implication rules (\S\ref{subsec:asymm}).

All models were implemented in TensorFlow~\cite{tensorflow2015}.
We use the hyperparameters of \newcite{riedel2013relation}, with $k=100$ hidden dimensions and a weight of $\alpha=0.01$ for the $L_2$ regularization loss.  
We use ADAM \cite{kingma2014adam} for optimization with an initial learning rate of 0.005 and a mini-batch size of $8192$. 
The embeddings are initialized by sampling uniformly from $[-0.1,0.1]$ and we use $\tilde{\beta}=0.1$ for the implication loss throughout our experiments.

\begin{table}[t!]
\centering
\resizebox{1.0\columnwidth}{!}{
    \begin{tabular}{ l l | c c c c c }
    \toprule
                  Test relation & \# & R13-F & F & FS & FSL \\
    \midrule
            person/company &  106 &     0.75 &   0.73 &   0.74 & {\bf 0.77} \\
      location/containedby &   73 &     0.69 &   0.62 &   0.70 & {\bf 0.71} \\
      person/nationality   &   28 &     0.19 &   0.20 &   0.20 & {\bf 0.21} \\
     author/works\_written &   27 &     0.65 &   {\bf   0.71} &   0.69 &   0.65 \\
      person/place\_of\_birth &   21 &  0.72 &  0.69 & {\bf   0.72} &   0.70 \\
              parent/child &   19 &   0.76 &   0.77 &   0.81 & {\bf   0.85} \\
      person/place\_of\_death &   19 &  0.83 &  {\bf 0.85} &   0.83 &   {\bf 0.85} \\
      neighborhood/neighborhood\_of &   11 &   {\bf 0.70} & 0.67 &   0.63 &   0.62 \\
       person/parents &    6 &   0.61 & 0.53 &   {\bf 0.66} &   {\bf 0.66} \\
       company/founders &    4 &   {\bf 0.77} & 0.73 &   0.64 &   0.67 \\
       sports\_team/league &    4 &  {\bf 0.59} &  0.44 &   0.43 &   0.56 \\
      team\_owner/teams\_owned &    2 &   0.38 &  {\bf 0.64} &   {\bf 0.64} &   0.61 \\
      team/arena\_stadium &    2 &  {\bf 0.13} &   {\bf 0.13} &   {\bf 0.13} &   0.12 \\
       film/directed\_by &    2 &  {\bf 0.50} &   0.18 &   0.17 &   0.13 \\
      broadcast/area\_served &    2 & 0.58 & 0.83 &   0.83 & {\bf   1.00} \\
      structure/architect &    2 & {\bf   1.00} & {\bf   1.00} & {\bf   1.00} & {\bf   1.00} \\
      composer/compositions &    2 &  {\bf 0.67} &  0.64 &   0.51 &   0.50 \\
       person/religion &    1 &  {\bf   1.00} & {\bf   1.00} & {\bf   1.00} & {\bf   1.00} \\
      film/produced\_by &    1 &  0.50 & {\bf   1.00} & {\bf   1.00} &   0.33 \\
    \midrule
              Weighted MAP &      &   0.67 &  0.65 &   0.67 &   0.69 \\
    \bottomrule          
    \end{tabular}
}
\caption{Weighted mean average precision for our reimplementation of the matrix factorization model (F) compared to restricting the entity-pair space (FS) and injecting WordNet rules (FSL). Model F results by Riedel et al. (2013) are denoted as R13-F.}
\label{tab:results}
\end{table}

\subsection{Restricted Embedding Space}\label{subsec:Frestricted}

\begin{table*}[t!]
    \centering
    \caption{Top patterns for a randomly sampled dimension in non-restricted and restricted embedding space .}
    \resizebox{0.9\textwidth}{!}{
    \begin{tabular}{ll}
        \toprule
        Model F (non-restricted) & Model FS (restricted)\\
        \midrule
        {\tt nsubj<-represent->dobj} & {\tt rcmod->return->prep->to->pobj}\\
        {\tt appos->member->prep->of->pobj->team->nn} & {\tt nn<-return->prep->to->pobj}\\
        {\tt nsubj<-die->dobj} & {\tt nsubj<-return->prep->to->pobj}\\
        {\tt nsubj<-speak->prep->about->pobj} & {\tt rcmod->leave->dobj}\\
        {\tt appos->champion->poss} & {\tt nsubj<-quit->dobj}\\
        \bottomrule
    \end{tabular}
    }
    \label{tab:toprelations}
\end{table*}

Before incorporating external commonsense knowledge into relation representations, we were curious how much we lose by restricting the entity-tuple space to approximately Boolean embeddings.
We evaluate our models on the New York Times dataset introduced by \newcite{riedel2013relation}.
Surprisingly, we find that the expressiveness of the model does not suffer from this strong restriction. 
From Table~\ref{tab:results} we see that restricting the tuple-embedding space seems to perform slightly better (FS) as opposed to a real-valued tuple-embedding space (F), suggesting that this restriction has a regularization effect that improves generalization. 
We also provide the original results for model F by \newcite{riedel2013relation} (denoted as R13-F) for comparison. Due to a different implementation and optimization procedure, the results for our model F and R13-F are not identical.

Inspecting the top relations for a sampled dimension in the embedding space reveals that the relation space of model FS more closely resembles clusters than that of model F (Table~\ref{tab:toprelations}). 
We hypothesize that this might be caused by approximately Boolean entity-tuple representations in model FS, resulting in attribute-like entity-tuple vectors that capture which relation clusters they belong to.

\subsection{Zero-shot Learning}\label{subsec:zeroshot}
The zero-shot learning experiment performed in \newcite{rocktaschel2015injecting} leads to an important finding: when injecting implications with right-hand sides for Freebase relations for which no or very limited training facts are available, the model should be able to infer the validity of Freebase facts for those relations based on rules and correlations between textual surface patterns. 

We inject the same hand-picked relations as used by \newcite{rocktaschel2015injecting}, after removing all Freebase training facts. 
The lifted rule injection (model FSL) reaches a weighted MAP of $0.35$, comparable with $0.38$ by the Joint model from \newcite{rocktaschel2015injecting} (denoted R15-Joint). 
Note that for this experiment we initialized the Freebase relations implied by the rules with negative random vectors (sampled uniformly from $[-7.9, -8.1]$). The reason is that without any negative training facts for these relations, their components can only go up due to the implication loss, and we do not want to get values that are too high before optimization.  

\begin{figure}[t!]
\centering
\includegraphics[width=1.1\columnwidth]{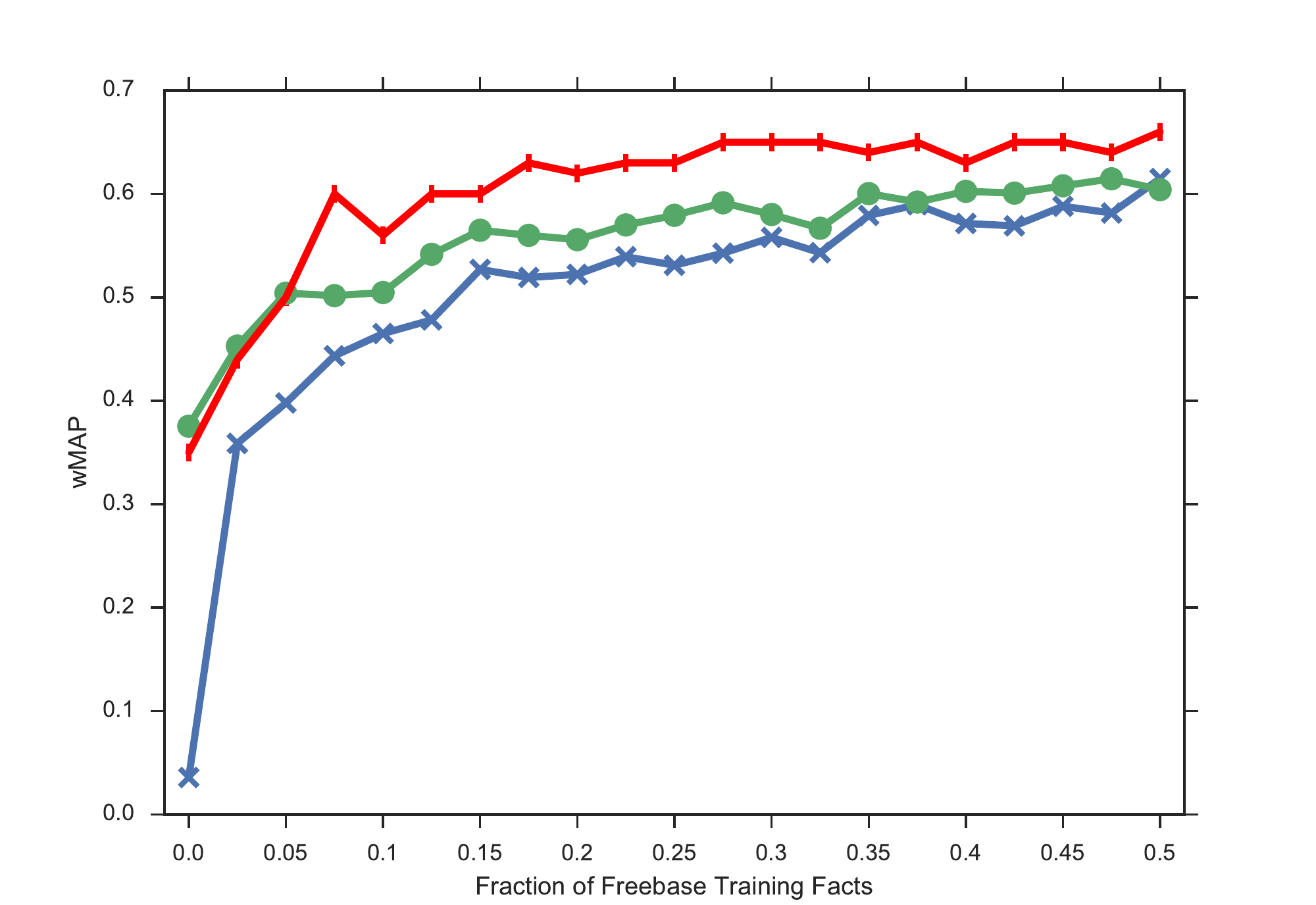}
\caption{Weighted MAP for injecting hand-picked rules as a function of the fraction of Freebase training facts. Comparison between model F (lowest, in blue), R15-Joint (middle, in green) and model FSL (highest, in red).}
\label{fig:zeroshot}
\end{figure}

Figure~\ref{fig:zeroshot} shows how the relation extraction performance improves when more Freebase relation training facts are added. 
It effictively measures how well the proposed models, matrix factorization (F), propositionalized rule injection (R15-Joint), and our model (FSL), can make use of the provided rules and correlations between textual surface form patterns and increased fractions of Freebase training facts.
Although FSL starts at a lower performance than R15-Joint when no Freebase training facts are present, it outperforms R15-Joint and a plain matrix factorization model by a substantial margin when provided with more than $7.5\%$ of Freebase training facts.
This indicates that, in addition to being much faster than R15-Joint, it can make better use of provided rules and few training facts. We attribute this to the Bayesian personalized ranking loss instead of the logistic loss used in \newcite{rocktaschel2015injecting}. The former is compatible with our rule-injection method, but not with the approach of maximizing the expectation of propositional rules used by R15-Joint.

\subsection{Injecting Knowledge from WordNet}\label{subsec:injectWordNet}
The main purpose of this work is to be able to incorporate rules from external resources for aiding relation extraction.
We use WordNet hypernyms to generate rules for the NYT dataset. 
To this end we iterate over all surface form patterns in the dataset and attempt to replace words in the pattern by their hypernyms. If the resulting pattern is contained in the dataset, we generate the corresponding rule. For instance, we generate a rule \verb~appos->diplomat->amod~ $\Rightarrow$ \verb~appos->official->amod~ since both patterns are contained in the NYT dataset and we know from WordNet that a diplomat is an official. 
This leads to $427$ rules from WordNet that we subsequently annotate manually to obtain $36$ high-quality rules.
Note that none of these rules directly imply a Freebase relation. Although the test relations all originate from Freebase, we still hope to see improvements by transitive effects, \ie{} better surface form representations that in turn help to predict Freebase facts.

We show results obtained by injecting these WordNet rules in Table~\ref{tab:results} (column FSL). The weighted MAP measure increases by 2\% with respect to model FS, and 4\% compared to our reimplementation of the matrix factorization model F.
This demonstrates that imposing a partial ordering based on implication rules can be used to incorporate logical commonsense knowledge and increase the quality of information extraction systems.
Note that our evaluation setting guarantees that only indirect effects of the rules are measured, \ie{} we do not use any rules directly implying test relations.
This shows that injecting such rules influences the relation embedding space beyond only the relations explicitly stated in the rules. 
For example, injecting the rule \verb~appos<-father->appos~ $\Rightarrow$ \verb~poss<-parent->appos~ can contribute to improved predictions for the test relation \verb~parent/child~.

\subsection{Visualizing Relation Embeddings}\label{subsec:visualizing}
\begin{figure}[t!]
     \centering
     \subfloat[b][]{\includegraphics[width=0.5\columnwidth]{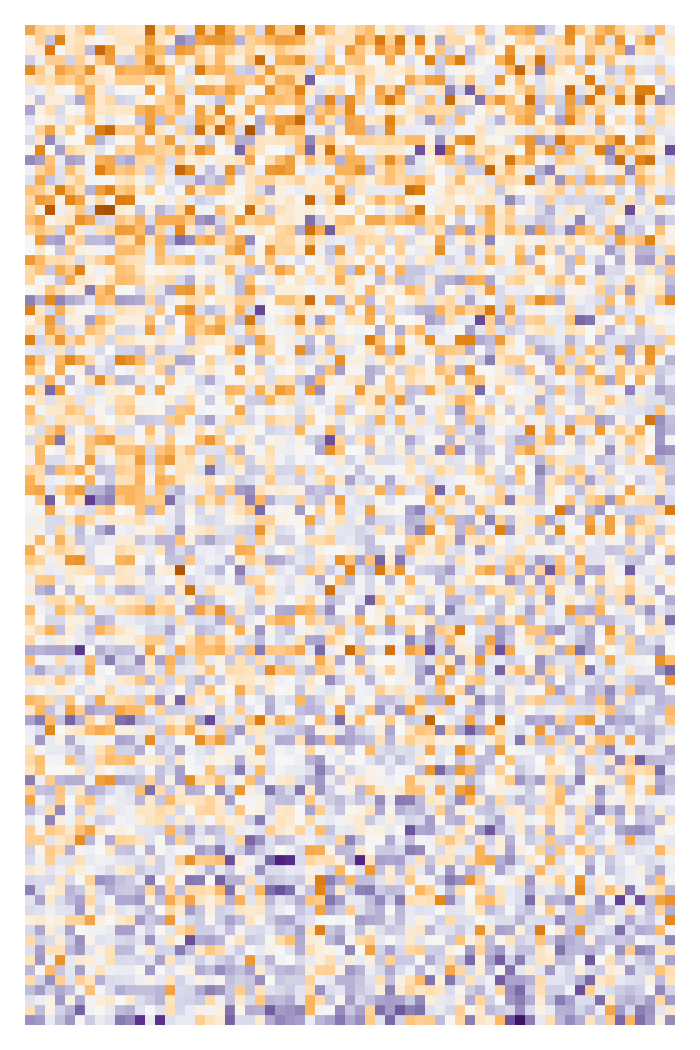}\label{fig:embeddings}}
     \hfill
     \subfloat[b][]{\includegraphics[width=0.5\columnwidth]{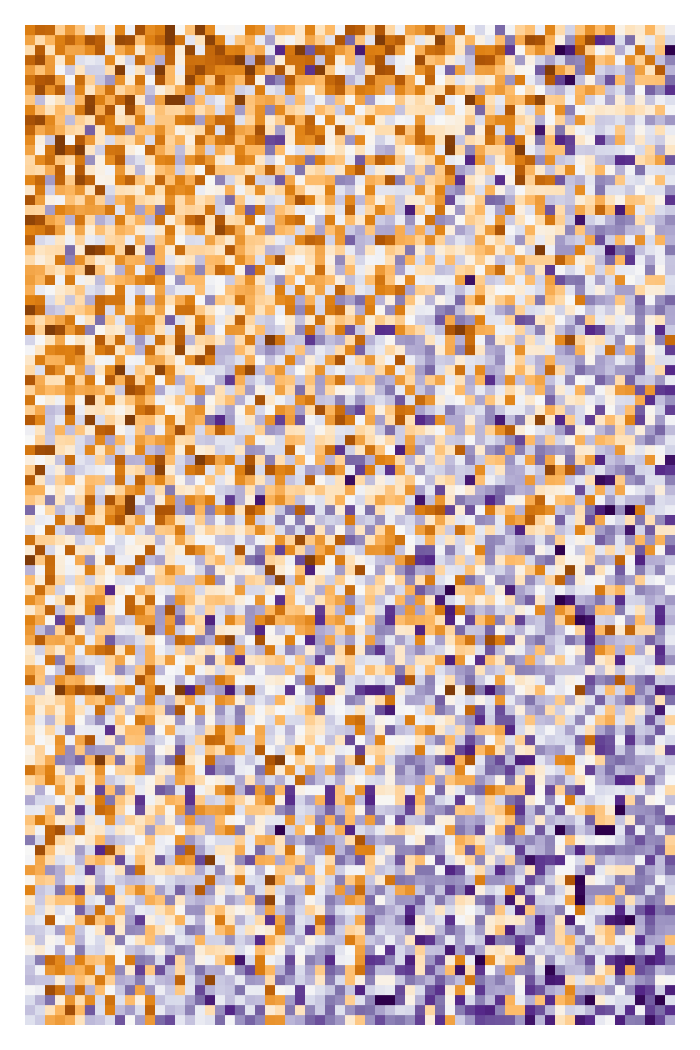}\label{fig:wordnet_embeddings}}
     \label{results}
     \caption{Visualization of embeddings (columns) for the relations that appear in the high-quality WordNet rules, (a) without and (b) with injection of these rules. Values range from -1 (orange) via 0 (white) to 1 (purple). Best viewed in color.}
\end{figure}

We provide a visual inspection of how the structure of the relation embedding space changes when rules are imposed. We select all relations involved in the WordNet rules, and gather them as columns in a single matrix, sorted by increasing $\ell_1$ norm (values in the $100$ dimensions are similarly sorted). Figures~\ref{fig:embeddings} and \ref{fig:wordnet_embeddings} show the difference between model F (without injected rules) and FSL (with rules). 
The values of the embeddings in model FSL are more polarized, \ie{} we observe stronger negative or positive components than for model F. 
Furthermore, FSL also reveals a clearer difference between the left-most (mostly negative, more specific) and right-most (predominantly positive, more general) embeddings (\ie{} a clearer separation between positive and negative values in the plot), which results from imposing the order relation in eq.~(\ref{eq:obj3_impl}) when injecting implications.

\subsection{Efficiency of Lifted Injection of Rules}\label{subsec:time}
In order to get an idea of the time efficiency of injecting rules, we measure the time per epoch when restricting the program execution to a single 2.4GHz CPU core. 
We measure on average 6.33s per epoch without rules (model FS), against 6.76s and 6.97s when injecting the 36 high-quality WordNet rules and the unfiltered 427 rules (model FSL), respectively.
Increasing the amount of injected rules from 36 to 427 leads to an increase of only 3\% in computation time, even though in our setup all rule losses are used in every training batch. 
This confirms the high efficiency of our lifted rule injection method.

\subsection{Asymmetric Character of Implications}\label{subsec:asymm}

\begin{table*}[t!]
\centering
\resizebox{1.9\columnwidth}{!}{
\begin{tabular}{ l c l | c c | c c}
\toprule
& rule &  & \multicolumn{2}{c|}{model FSL} & \multicolumn{2}{c}{model FS}\\
\multicolumn{1}{c}{$r_p$} & $\Rightarrow$ & \multicolumn{1}{c|}{$r_q$} & $\sigma(\dotre{r_q}{t_p})$ & $\sigma(\dotre{r_p}{t_q})$ & $\sigma(\dotre{r_q}{t_p})$ & $\sigma(\dotre{r_p}{t_q})$ \\
\midrule
{\tt appos->party->amod}  & $\Rightarrow$ & {\tt appos->organization->amod}  &  0.99  &  0.22  & 0.70  & 0.86  \\
{\tt poss<-father->appos}  & $\Rightarrow$ & {\tt poss<-parent->appos}  &  0.96  &  0.00  & 0.72 & 0.89  \\
{\tt appos->prosecutor->nn}  & $\Rightarrow$ & {\tt appos->lawyer->nn}  &  0.99  &  0.01  & 0.87 & 0.80  \\
{\tt appos->daily->amod}  & $\Rightarrow$ & {\tt appos->newspaper->amod}  &  0.98  &  0.79  & 0.90 & 0.86  \\
{\tt appos->ambassador->amod}  & $\Rightarrow$ & {\tt appos->diplomat->amod}  &  0.31  &  0.05  & 0.93 & 0.84  \\
\midrule
\multicolumn{3}{c|}{average over 36 high-quality Wordnet rules}    &  0.95  &  0.28  & 0.74 & 0.70  \\
\bottomrule          
\end{tabular}
}
\caption{Average of $\sigma(\dotre{r_q}{t})$ over all inferred facts $\fact{r_q}{t_p}$ for tuples $\emb{t_p}$ from training items for relation $r_p$, and vice versa, for Wordnet implications $\emb{r_p}\Rightarrow\emb{r_q}$, and model FSL (injected rules) vs.\ model FS (no rules).}
\label{tab:asymmetry}
\end{table*}

In order to demonstrate that injecting implications conserves their asymmetric nature, we perform the following experiment.
After incorporating high-quality Wordnet rules $r_p\Rightarrow r_q$ into model FSL we select all of the tuples $t_p$ that occur with relation $r_p$ in a training fact $\fact{r_p}{t_p}$. 
Matching these with relation $r_q$ should result in high values for the scores $\dotre{r_q}{t_p}$, if the implication holds. 
If however the tuples $t_q$ are selected from the training facts $\fact{r_q}{t_q}$, and matched with relation $r_p$, the scores $\dotre{r_p}{t_q}$ should be much lower if the inverse implication does not hold (in other words, if $r_q$ and $r_p$ are not equivalent). 
Table~\ref{tab:asymmetry} lists the averaged results for 5 example rules, and the average over all relations in WordNet rules, both for the case with injected rules (model FSL), and without rules (model FS).
For easier comparison, the scores are mapped to the unit interval via the sigmoid function. 
This quantity $\sigma(\dotre{r}{t})$ is often interpreted as the probability that the corresponding fact holds  \cite{riedel2013relation}, but because of the BPR-based training, only differences between scores play a role here. 
After injecting rules, the average scores of facts inferred by these rules (\ie{} column $\sigma(\dotre{r_q}{t_p})$ for model FSL) are always higher than for facts (incorrectly) inferred by the inverse rules (column $\sigma(\dotre{r_p}{t_q})$ for model FSL). 
In the fourth example, the inverse rule leads to high scores as well (on average 0.79, vs.\ 0.98 for the actual rule). This is due to the fact that the {\tt daily} and {\tt newspaper} relations are more or less equivalent, such that the components of $\emb{r_p}$ are not much below those of $\emb{r_q}$.
For the last example (the {\tt ambassador} $\Rightarrow$ {\tt diplomat} rule), the asymmetry in the implication is maintained, although the absolute scores are rather low for these two relations. 

The results for model~FS reflect how strongly the implications in either direction are latently present in the training data. 
We can only conclude that model~FS manages to capture the similarity between relations, but not the asymmetric character of implications. 
For example, purely based on the training data, it appears to be more likely that the {\tt parent} relation implies the {\tt father} relation, than vice versa. This again demonstrates the importance and added value of injecting external rules capturing commonsense knowledge.

\section{Conclusions}\label{sec:conclusion}

We presented a novel, fast approach for incorporating first-order implication rules into distributed representations of relations. 
We termed our approach  `lifted rule injection', as it avoids the costly grounding of first-order implication rules and is thus independent of the size of the domain of entities.
By construction, these rules are satisfied for any observed or unobserved fact.
The presented approach requires a restriction on the entity-tuple embedding space. 
However, experiments on a real-world dataset show that this does not impair the expressiveness of the learned representations. 
On the contrary, it appears to have a beneficial regularization effect. 

By incorporating rules generated from WordNet hypernyms, our model improved over a matrix factorization baseline for knowledge base completion. 
Especially for domains where annotation is costly and only small amounts of training facts are available, our approach provides a way to leverage external knowledge sources for inferring facts. 

In future work, we want to extend the proposed ideas beyond implications towards general first-order logic rules.
We believe that supporting conjunctions, disjunctions and negations would enable to debug and improve representation learning based knowledge base completion.
Furthermore, we want to integrate these ideas into neural methods beyond matrix factorization approaches.

\section*{Acknowledgments}
This work was supported by the Research
Foundation - Flanders (FWO), Ghent University -
iMinds, Microsoft Research through its PhD Scholarship
Programme, an Allen Distinguished Investigator
Award, and a Marie Curie Career Integration Award.


\bibliography{foil_EMNLP}
\bibliographystyle{emnlp2016}

\end{document}